\documentclass[cameraready]{Interspeech}

\title{Better Late Than Never: Meta-Evaluation of Latency Metrics\\ for Simultaneous Speech-to-Text Translation}

\author[affiliation={1}, orcid=0000-0003-2332-7666, ]{Peter}{Polák}
\author[affiliation={2}, orcid=0000-0002-4494-8886,]{Sara}{Papi}
\author[affiliation={2}, orcid=0000-0001-7480-2231,]{Luisa}{Bentivogli}
\author[affiliation={1}, orcid=0000-0002-0606-0050, ]{Ondřej}{Bojar}

\address{
    $^1$ Charles University, Czech Republic \\
    $^2$ Fondazione Bruno Kessler, Italy
}

\email{$^1$\texttt{\{polak,bojar\}@ufal.mff.cuni.cz}, $^2$\texttt{\{spapi,bentivo\}@fbk.eu}}


\usepackage{comment}

\newcommand{\mwerSegmenter}{\textsc{mwer\hspace{0pt}Seg\hspace{0pt}men\hspace{0pt}ter}\xspace}
\newcommand{\segmenter}{\textsc{Soft\hspace{0pt}Seg\hspace{0pt}men\hspace{0pt}ter}\xspace}

\newcommand{\toolkit}{\textsc{Omni\hspace{0pt}ST\hspace{0pt}Eval}\xspace}
\newcommand{\toolkiturl}{https://github.com/pe-trik/OmniSTEval}

\usepackage[noabbrev,capitalize]{cleveref}

\usepackage{multirow}
\usepackage{booktabs}

\usepackage{twemojis}

\usepackage{graphicx}
\usepackage{subcaption}

\newcommand{\Long}[1]{$\substack{\text{Long}\\\text{#1}}$}
\newcommand{\Stream}[1]{$\substack{\text{Stream}\\\text{#1}}$}

\usepackage[acronym, nomain, nopostdot]{glossaries}

  \newacronym{mt}{MT}{Machine Translation}
  \newacronym{asr}{ASR}{Automatic Speech Recognition}
  \newacronym{vad}{VAD}{Voice Activity Detection}
  \newacronym{sst}{SimulST}{Simultaneous Speech Translation}
  \newacronym{st}{ST}{Speech Translation}
  \newacronym{la}{LA}{Local Agreement}
  \newacronym{sp}{SP}{Shared Prefix}
  \newacronym{lan}{LA-$n$}{Generalized Local Agreement}
  \newacronym{spn}{SP-$n$}{Generalized Shared Prefix}
  \newacronym{hn}{hold-$n$}{Hold-$n$}

  \newacronym{bwbs}{BWBS}{Blockwise Beam Search}
  \newacronym{ibwbs}{IBWBS}{Incremental Blockwise Beam Search}
  \newacronym{bs}{BS}{Beam Search}
  \newacronym{svo}{SVO}{Subject-Verb-Object}
  \newacronym{sov}{SOV}{Subject-Object-Verb}

  \newacronym{rtf}{RTF}{Real-Time Factor}
  \newacronym{al}{AL}{Average Lagging}

  \newacronym{e2e}{E2E}{End-to-End}
  \newacronym{ctc}{CTC}{Connectionist Temporal Classification}

  \newacronym{aed}{AED}{Attention-Based Encoder-Decoder}
  \newacronym{st-ctc}{ST-CTC}{CTC-Augmented AED for ST}

  \newacronym{rnn}{RNN}{Recurrent Neural Network}
  \newacronym{rnnt}{RNN-T}{Recurrent Neural Network Transducer}
  \newacronym{tts}{TTS}{Text-To-Speech}
  
  \newacronym{mfc}{MFC}{Montreal Forced Aligner}

\usepackage{amsmath,amsfonts,amsthm,bm}

\renewcommand{\paragraph}[1]{\textbf{#1}}
\newcommand{\citealp}[1]{\cite{#1}}
\newcommand{\citep}[1]{\cite{#1}}
\newcommand{\citet}[1]{\cite{#1}}

\usepackage[dvipsnames]{xcolor}

\begin{document}

\maketitle

\begin{abstract}


Simultaneous speech-to-text translation systems must balance translation quality with latency. Although quality evaluation is well established, latency measurement remains a challenge. Existing metrics produce inconsistent results, especially in short-form settings with artificial presegmentation. We present the first comprehensive meta-evaluation of latency metrics across language pairs and systems. We uncover a structural bias in current metrics related to segmentation. We introduce YAAL (Yet Another Average Lagging) for a more accurate short-form evaluation and LongYAAL for unsegmented audio. We propose \segmenter{}, a resegmentation tool based on soft word-level alignment. We show that YAAL and LongYAAL, together with \segmenter{}, outperform popular latency metrics, enabling more reliable assessments of short- and long-form simultaneous speech translation systems.
We implement all artifacts within the \toolkit{} toolkit: \url{\toolkiturl}.

\end{abstract}

\section{Introduction}
\label{sec:better-late}

\begin{figure}[t]
    \centering
    \includegraphics[width=\linewidth]{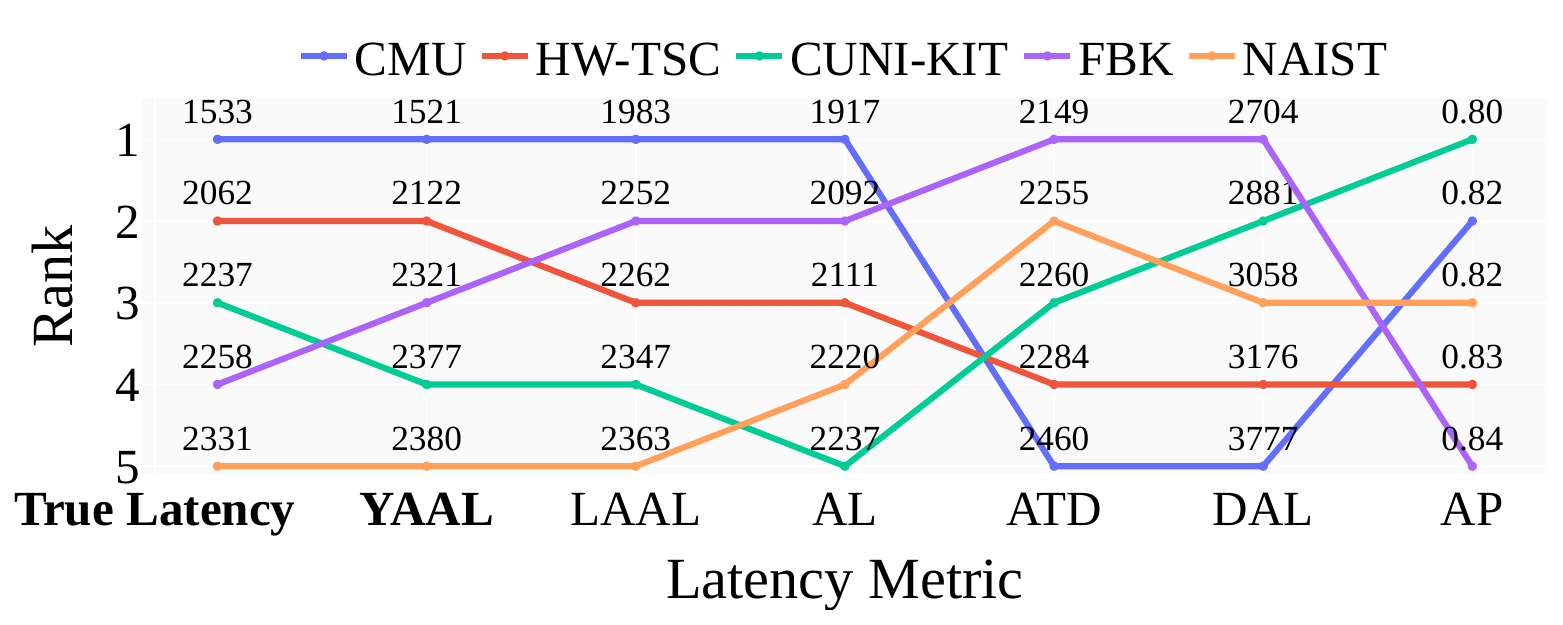}
    \caption{Ranking of the systems submitted to the IWSLT 2023 Simultaneous Speech Translation Track according to the True Latency, the proposed automatic metric YAAL, and the official five latency metrics.}
    \label{fig:iwslt23-ranks}
\end{figure}

Although translation quality metrics are extensively studied in offline \gls{st} and in the related field of machine translation \cite{freitag-etal-2022-results,freitag-etal-2023-results,zouhar-etal-2024-pitfalls}, very little attention has been paid to latency metrics in \gls{sst}.
While the most common latency metrics in \gls{sst}~\cite{cho2016neural,ma-etal-2019-stacl,cherry2019thinking,polak-etal-2022-cuni,papi-etal-2022-generation,kano23_interspeech} rely on different approximations, they all base their calculations on simplifying assumptions, such as uniform word duration, the absence of pauses, and strict monotonic alignment between source speech and target translation.
However, despite relying on the same assumptions, these metrics often produce very inconsistent assessments. 
This inconsistency is clearly illustrated in the results of the IWSLT 2023 Shared Task on Simultaneous Translation~\cite{agrawal-etal-2023-findings}, where different metrics produced substantially different rankings (see \cref{fig:iwslt23-ranks}).
This variability raises serious concerns about the validity of current evaluation protocols and their ability to support meaningful comparisons between systems.
Moreover, this risk can be further exacerbated when shifting from dealing with already pre-segmented speech, i.e., \textit{short-form} \gls{sst}, to unsegmented audio, i.e., \textit{long-form} \gls{sst}, where information about sentence boundaries is not available, further complicating the evaluation~\cite{papi2025real}. 
Although \cite{iranzo-sanchez-etal-2025-going} showed that reporting a single latency value in a short-form setting may lead to misleading comparisons, the underlying reasons for inconsistencies between latency metrics and their overall reliability remain unclear.

In this paper, we present the first comprehensive meta-evaluation of latency metrics for \gls{sst} in several aspects, including diverse systems, language pairs, and short- and long-form regimes.
Through an in-depth analysis of systems submitted to recent IWSLT Simultaneous Speech Translation Shared Tasks~\cite{anastasopoulos-etal-2022-findings,agrawal-etal-2023-findings,ahmad-etal-2024-findings}, we reveal that existing metrics can lead to misleading conclusions and hinder effective system design.
We show that the inconsistent evaluations are not primarily due to the aforementioned assumptions, but rather to a structural bias in how latency is measured---particularly in how segmentation influences \gls{sst} models' behavior.
Motivated by these findings, we propose YAAL (Yet Another Average Lagging), a refined latency metric designed to mitigate the biases present in existing latency metrics.
Using the proposed YAAL metric, we identify a ``degenerate'' behavior in some systems, where the majority of the translation occurs after the input ends.
We introduce a diagnostic test to detect this behavior, enabling practitioners to identify when a system exhibits degeneracy and when latency metrics may be unreliable.
Furthermore, we also show that resegmentation, which pairs segment-level predictions with their corresponding reference, is necessary to produce meaningful latency measurements for long-form \gls{sst}.
To this end, we introduce \segmenter{}, a new resegmentation tool, and extend our YAAL to LongYAAL, which deals with audio streams.
Compared to the current alignment tool used in the speech translation community \cite{matusov-etal-2005-evaluating}, \segmenter{} significantly improves alignment quality, enabling more accurate evaluation in long-form scenarios.
The code for YAAL, its LongYAAL long-form variant, and \segmenter{} is available within the \toolkit{} toolkit.\footnote{\toolkiturl}
\section{Background}

In the following, we describe the metrics currently used for both the short-form (\S\ref{subsec:short-latency-metrics}) and long-form (\S\ref{subsec:long-latency-metrics}) regimes.
Throughout the paper, we assume incremental SimulST systems, i.e., systems that cannot revise their outputs because they are not affected by flickering problems, which are leading current research efforts on the topic \citep{papi2025real}.

\subsection{Short-Form SimulST Latency Metrics}
\label{subsec:short-latency-metrics}

The short-form is the most common evaluation regime of SimulST \cite{anastasopoulos-etal-2022-findings,agrawal-etal-2023-findings,ahmad-etal-2024-findings}, where all recordings of the test set are divided, usually following sentence boundaries, into short segments of a few seconds. 
Each segment consists of source audio $\mathbf{X} = [x_1, \dots,x_{|\mathbf{X}|}]$, where $x_i$ is a small portion of raw audio, i.e., the audio chunk, hypothesis translation $\mathbf{Y} = [y_1, \dots,y_{|\mathbf{Y}|}]$, and reference translation $\mathbf{Y^R} = [y^R_1, \dots,y^R_{|\mathbf{Y^R}|}]$. 
Each audio chunk $x_i$ is fed incrementally to the system, 
which concurrently
outputs a translation token $y_{j}$ at timestamp $d_j$, i.e., the total duration of audio chunks up to and including the audio chunk $x_i$.  %
In these settings, we describe the latency metrics operating in the short-form regime.

\paragraph{Average Proportion (AP; {\rm\citealp{cho2016neural}})} measures the average proportion of input speech read when emitting a target token:

\begin{equation}
    \text{AP} = \frac{1}{|\mathbf{X}||\mathbf{Y}|} \sum_{i=1}^{|\mathbf{Y}|} d_i.
    \label{eq:ap}
\end{equation}

\paragraph{Average Lagging (AL; {\rm\citealp{ma-etal-2019-stacl}})} for simultaneous machine translation and modified for speech by \citet{ma-etal-2020-simuleval} defines the latency as the average delay behind an ideal policy:

\begin{equation}
    \text{AL} = \frac{1}{\tau(\mathbf{X})} \sum_{i=1}^{\tau(\mathbf{X})} d_i - d^*_i,%
    \label{eq:al}
\end{equation}

\noindent where $\tau(\mathbf{X}) = \text{min}\{i | d_i=|\mathbf{X}||\}$ is the index of the hypothesis token when the model reaches the end of the source sentence, also known as the cutoff point. AL considers delays up to and including the one associated with the token at the cutoff point.
The $i$-th delay of the ideal policy is defined as $d^*_i = (i-1)/\gamma$, where $\gamma = |\mathbf{Y^R}|/|\mathbf{X}|$.

\paragraph{Length-Aware Average Lagging (LAAL; \rm\citealp{polak-etal-2022-cuni,papi-etal-2022-generation})} 
is an AL modification that is robust to overgeneration, i.e., when the hypothesis $\mathbf{Y}$ is much longer than $\mathbf{Y^R}$, which makes the original AL produce negative delays when $|\mathbf{Y}| \gg |\mathbf{Y^R}|$. 
To overcome this problem, which was unduly rewarding overgenerating systems,
LAAL modifies the $\gamma$ as follows:
$\gamma = \max(|\mathbf{Y}|,|\mathbf{Y^R}|)/|\mathbf{X}|$.

\paragraph{Differentiable Average Lagging (DAL; {\rm\citealp{cherry2019thinking}})} %
modifies AL by introducing
a minimal delay of $1/\gamma$ after each
step.
Unlike AL and LAAL, DAL considers all tokens in the hypothesis, 
without
cutoff after $i > \tau(\mathbf{X})$,
and instead DAL penalizes each write operation by at least $1/\gamma$: $d'_i = \max(d_i, d'_{i-1} + 1/\gamma)$, where $d'_1 = d_1$.
%
%
%
%
%
%
%

%

\paragraph{Average Token Delay (ATD; {\rm\citealp{kano23_interspeech}})} assumes that the source speech, similar to the translation, consists of discrete tokens. ATD defines a fixed duration for speech tokens of 300ms
and divides the input speech and translation into $C$ chunks, where the $c$-th translation chunk $y^c$ is translated conditioned on the source chunk $x^c$ and previous translation chunks $y^1,\dots,y^{c-1}$. ATD is then defined as the average delay between each translation and the corresponding source tokens:
\begin{equation}
    \text{ATD} = \frac{1}{|\mathbf{Y}|} \sum_{i=1}^{|\mathbf{Y}|} (T(y_t) - T(x_{a(t)})),
    \label{eq:atd}
\end{equation}
where $T(\cdot)$ is the end time of the source/translation token and
\begin{equation}
    a(t) = \begin{cases}
        s(t), & \text{if} s(t) \leq L_{acc}(x^{c(t)}) \\
        L_{acc}(x^{c(t)}), & otherwise,
    \end{cases}
\end{equation}
is an index of a source token corresponding to translation token $y_t$, where $L_{acc}(x^c)$ is the number of source tokens in the chunk $x^c$ and $s(t) = t - max(0,L_{acc}(y^{c(t)-1})-L_{acc}(x^{c(t)-1}))$
handles the case where more tokens are generated than read, i.e., $y_t$ is aligned with $x_{t'}$, $t' < t$.

\subsection{Long-Form SimulST Latency Metrics}
\label{subsec:long-latency-metrics}

The long-form evaluation regime evaluates SimulST systems more realistically \cite{papi2025real}, as it assesses their ability to handle long audio streams, often spanning several minutes.
Since all metrics were developed for the short-form regime, 
recent studies \cite{papi2024streamatt,polak2024longformslt} 
resorted to resegmentation of translations and delays based on the reference translation \cite{matusov-etal-2005-evaluating}, and computed the metrics on the segment level. We explain the long-form variant of LAAL \cite{papi2024streamatt} below.

\paragraph{Streaming LAAL (StreamLAAL; {\rm\cite{papi2024streamatt}})} extends the LAAL metric to unsegmented audio streams $\mathbf{S}=[\mathbf{X}_1,...,\mathbf{X}_{|\mathbf{S}|}]$, paired with a continuous stream of predicted translations $\mathbf{Y_S}$. Since reference translations  $\mathbf{Y^R_{1}},...,\mathbf{Y^R_{|\mathbf{S}|}}$ are only available at segment-level $\mathbf{X_1},...,\mathbf{X}_{|\mathbf{S}|}$, prediction $\mathbf{Y_S}=[\mathbf{Y_{1}}, ..., \mathbf{Y_{|\mathbf{S}|}}]$ with the corresponding delays is segmented based on reference sentences $\mathbf{Y^R_s}$ to obtain segment-level predictions. Then, StreamLAAL is computed as:
\begin{equation}
\begin{split}
\label{equation:StreamLAAL}
{\substack{\text{Stream} \\ \text{LAAL}}}&=\frac{1}{|\mathbf{S}|} \sum_{s=1}^{|S|} \frac{1}{\tau(\mathbf{X_s})}\sum_{i=1}^{\tau(\mathbf{X_s})}d_i - d^*_i \\
\end{split}
\end{equation}
Where $d^*_i = (i-1) \cdot |\mathbf{X_s}| /{\max\{|\mathbf{Y_{s}}|,|\mathbf{Y^R_{s}|\}}}$
In practice, the LAAL metric is calculated for every speech segment $\mathbf{X_s}$ of the stream $\mathbf{S}$ and its corresponding reference $\mathbf{Y^R_{s}}$ with the automatically aligned prediction $\mathbf{Y_{s}}$ and then averaged over all the speech segments of the stream $\mathbf{X_1},...,\mathbf{X_{|\mathbf{S}|}}$. 

Alternatively, \citet{huber-etal-2023-end} proposed an evaluation with resegmentation provided by the system. The evaluation framework expects that the system outputs the segment's start and end timestamps that align with the source. First, most systems, including all the IWSLT systems, do not output this information, and relying on the system's self-reported alignment might hinder the reliability. Second, as we empirically show in \S\ref{subsec:short-eval}, relying on potentially low-quality resegmentation significantly lowers the accuracy of latency evaluation. Finally, different segmentations render the observed latencies incomparable across different systems. 
\section{Overcoming the Pitfalls in SimulST Latency Metrics}
\label{sec:pitfalls}

\subsection{The Short-Form Regime}
\label{subsec:short}

\begin{figure*}[t]
    \centering
    \includegraphics[width=\linewidth]{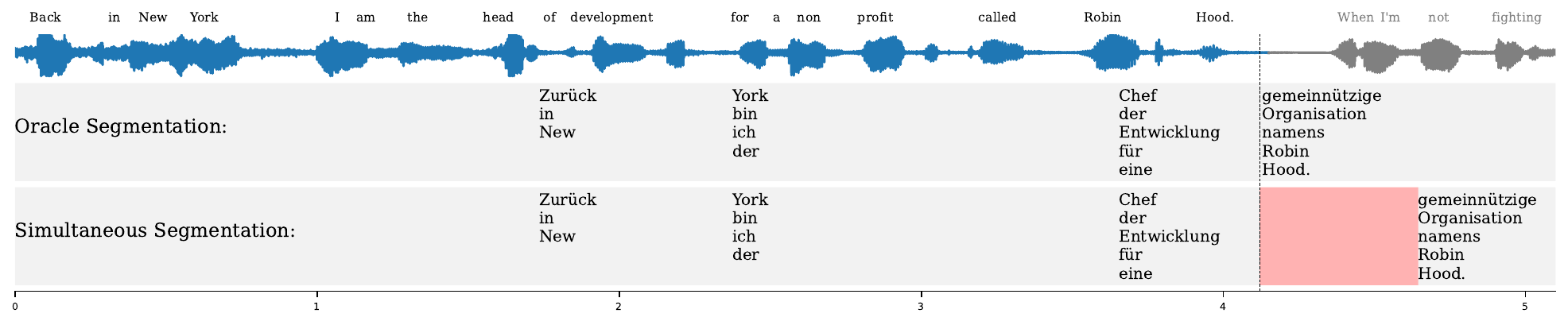}
    \caption{Translations and emission times of a \gls{sst} model.
    Words in a column were emitted at once, last five words or \textit{tail words} ``\textit{gemeinnützige Organisation namens Robin Hood.}'' depend on the segmentation: 
     \textbf{Oracle Segmentation}: Known beforehand. Once the model consumes the entire sentence, it is asked to finish the translation without additional delay. 
     \textbf{Simultaneous Segmentation}: The evaluation uses an online segmenter that needs extra time (the \textcolor{Red}{red area}: approximately 0.5s) to decide when the sentence ends. 
     }
    \label{fig:segmented-inference-example}
\end{figure*}

The use of audio segmentation in short-form evaluations significantly affects translation behavior and latency \cite{sperber-etal-2024-evaluating}.
In practice, short-form \gls{sst} systems are evaluated in a simulated environment where each segment is processed independently \cite{ma-etal-2020-simuleval}.
When the entire source segment has been consumed by the system, the translation is often still in progress.
At that point, the simulator requests the remaining translation, which the model emits without any additional delay.
This setup introduces two unrealistic conditions.
First, the audio is typically segmented in advance by a human annotator or an automatic model with access to the full audio (\textit{Oracle Segmentation}).
Second, the model is allowed to generate the remaining translation (hereinafter, \textit{tail words}) instantaneously once the input segment ends.
These factors unduly distort short-form evaluations, both by providing high-quality segmentation and eliminating the delay that would occur in a realistic setting, where the system must wait to confirm that the sentence has ended.
In a more realistic scenario, a model has to rely on online segmentation (\textit{Simultaneous Segmentation}) and thus delay the final translation until it is confident that the input sentence is complete, thereby introducing extra latency.
This discrepancy is illustrated in \cref{fig:segmented-inference-example}.

Based on these observations, we categorize the existing short-form latency metrics AL, LAAL, DAL, AP, and ATD into two groups, depending on whether they compute latency over all translated words or only a subset.
The first group (AP, DAL, and ATD) includes all translated words in the calculation. 
DAL attempts to mitigate the impact of tail words by adding a minimum delay of $1/\gamma$ after each generated word (also within the same step), thus ``spreading'' the tail beyond the sentence. 
However, $1/\gamma$ simply reflects the average source-to-target length ratio and does not accurately capture the system behavior for both the simultaneous words emitted at once and the tail words in settings without segmentation.
If multiple words are emitted as tail words, DAL can significantly overestimate latency. 
In the edge case of a system that waits for an end-of-segment signal before emitting any words (i.e., an offline system), the DAL latency value is then equal to the segment length,
failing to capture the system's true behavior---in this case, \textit{undefined} latency.
AP assigns a delay of 1 to each tail word, as the entire recording has to be processed to emit that word; thus, the proportion is 1.
Although AP correctly assigns a final latency of 1 to an offline system (where all words are tail words), it remains sensitive to segmentation, like DAL, because it treats all tail words equally, regardless of how many are emitted.
ATD also considers all translated words, including tail words.
However, unlike DAL, it does not apply corrections for tail word behavior, making it the most sensitive to segmentation artifacts among the three metrics.
The second group (AL and LAAL) computes the latency only for words emitted up to and including the cutoff point $\tau(\mathbf{X})$, which marks the first word generated after the end of the input segment.
This corresponds to the word ``\textit{gemeinnützige}'' in \cref{fig:segmented-inference-example}. 
As discussed, in the short-form regime with oracle segmentation, the $\tau(\mathbf{X})$-th and following words are often translated earlier than in a more realistic long-form scenario.
As a result, this cutoff introduces a systematic bias in the latency estimate, which may lead to either underestimation or overestimation, depending on the system's policy.

AP, DAL, ATD, AL and, more recently, LAAL have become established metrics in the short-form evaluation of \gls{sst}.
However, as discussed above, including any of the tail words in the latency computation leads to a systematic bias that undermines fair comparisons. 
To cope with this bias, we propose a new metric derived from the LAAL metric.

\subsubsection{Yet Another Average Latency (YAAL)} 
\label{sec:yaal-def}
We refine the LAAL formulation to better isolate the portion of the output that is actually produced in simultaneous settings, and we dub the new metric \emph{Yet Another Average Latency} (YAAL).
Specifically, we define a new cutoff point
\begin{equation}
    \tau_{\text{YAAL}}({X}) = \max_{i} \{i~|~d_i<|\mathbf{X}|\},
\end{equation}
which includes only words generated \emph{strictly before the end} of the input stream.
This corresponds to words up to and including ``\textit{eine}'' in \cref{fig:segmented-inference-example}, thereby avoiding distortion from tail words and yielding a more reliable latency estimate that remains consistent across different segmentation regimes.
As we show in~\S\ref{subsec:short-eval}, YAAL provides a more robust latency estimate that better reflects the true behavior of the system in simultaneous settings.

\subsubsection{The Long-Form Regime}
\label{subsec:long}

The long-form regime offers a more realistic evaluation setting by assessing systems on continuous, unsegmented audio streams that better reflect real-world use cases.
However, the widely used latency metrics were originally designed for the short-form regime and do not extend directly to this setting.

First, metrics such as AL, LAAL, and DAL are based on the $\gamma$ parameter, which represents the average length ratio of the target-to the source.
However, $\gamma$ can vary substantially between different segments within the same audio, leading to inconsistent latency estimates~\cite{iranzo-sanchez-etal-2021-stream-level}. 
Second, AP tends to converge towards 0.5 for long recordings \cite{ma-etal-2020-simuleval}.
Finally, ATD assumes that each speech token has a fixed duration and that source and target tokens align monotonically---assumptions that are overly restrictive and especially unrealistic for long-form speech.
To address these challenges, StreamLAAL has introduced the re-segmentation of long inputs into short segments and is computing latency on these units.
Although StreamLAAL provides an adaptation of existing metrics to long-form input, it relies on the \mwerSegmenter{} tool~\cite{matusov-etal-2005-evaluating}, which may introduce alignment errors~\cite{amrhein-haddow-2022-dont,polak2024longformslt}.
When coupled with LAAL, it also computes latency up to the cutoff word $\tau(\mathbf{X_i})$ (\cref{equation:StreamLAAL}), which can lead to systematic bias, as discussed in \S\ref{subsec:short}.
To overcome these limitations, we propose a new resegmentation method and an extension of the YAAL metric for the long-form regime.

\subsubsection{\segmenter{}}

We introduce a new resegmentation method inspired by \cite{polak2024longformslt}, employing a softer alignment strategy to more accurately match the translation output with the reference segments. 
We start by lowercasing and tokenizing both the reference and the system hypotheses.
This allows for a more precise alignment around the sentence ends, especially in cases where the reference and the model differ in sentence segmentation. 
Although we change the text for alignment, we keep the original version to ensure that the translation quality score remains accurate.
Additionally, we maintain the delay associated with each token and use it during the alignment process to prevent the alignment of tokens with future segments, which would lead to spurious negative latencies.
For alignment, we maximize the following score:

\begin{equation}
    \mathcal{S}(t_r,t_h) = \begin{cases}
        -\infty & s_r \geq d_h, \\
        -\infty & P(t_r) \oplus P(r_h),  \\
        \mathcal{S}_{\text{char}}(t_r,t_h) & \text{otherwise,}
    \end{cases}
\end{equation}
where the first branch of the equation prevents the alignment of tokens to future reference segments, where $s_r$ is the start of the reference segment and $d_h$ is the emission time of the hypothesis token.
The second branch prevents the alignment of punctuation tokens to non-punctuation tokens, where $t_r$ and $t_h$ are the reference and hypothesis tokens, $P(\cdot)$ is a function that indicates if the token is punctuation, and $\bigoplus$ is exclusive or.
Finally, the last branch computes character-level similarity as $\mathcal{S}_{\text{char}}(t_r,t_h) = (t_r \cap t_h) / (t_r \cup t_h)$.
In the case of character-based languages, such as Chinese, this reduces to an exact match.

\subsubsection{Long-Form YAAL (LongYAAL)}
We also extend YAAL to the long-form regime---i.e., LongYAAL.
Unlike StreamLAAL, LongYAAL includes all words in the latency computation, even those generated beyond the aligned segment boundaries $\mathbf{X_s}$, i.e., all $d_i$ for $i > \tau(\mathbf{X_s})$.
However, we exclude the final tail words produced after the end of the full stream $\mathbf{S}$, i.e., $d_i$ for $i > \tau(\sum_{s=1}^{|\mathbf{S}|}|\mathbf{X_s}|) = \tau(\mathbf{X})$.
This ensures that we include all words emitted beyond the segment boundaries $\mathbf{X_s}$, which are generated in a realistic simultaneous manner in the long-form setting,
but we do not include the tail words generated at the end of the entire stream $\mathbf{S}$, which would introduce bias as discussed in \S\ref{subsec:short}.
Formally, LongYAAL is defined as
\begin{equation}
    \text{LongYAAL} = \frac{1}{\tau(\mathbf{S})} \sum_{s=1}^{\tau(\mathbf{S})}  \frac{1}{\tau(\mathbf{X})} \sum_{i=1}^{\tau(\mathbf{X})} d_i - d^*_i,
\end{equation}
where $\tau(\mathbf{S})$ is the number of segments that have at least one word generated before the end of the entire stream, and $\tau({\mathbf{X}})$ is the cutoff point that allows only words generated before the end of the entire stream $\mathbf{S}$.
If the stream $\mathbf{S}$ consists of a single segment, LongYAAL coincides with YAAL.
\section{Experimental Settings}
\label{sec:exp}

\subsection{Data} 

For the short-form regime, we use systems from the IWSLT 2022 and 2023 Simultaneous Speech Translation tracks~\cite{anastasopoulos-etal-2022-findings,agrawal-etal-2023-findings}, along with logs from the MuST-C \texttt{tst-COMMON} test set~\cite{CATTONI2021101155}.
For the long-form regime, we use IWSLT 2025 logs: for EN-to-\{DE, ZH, JA\}, we evaluate on the ACL 60/60 dataset~\cite{salesky-etal-2023-evaluating} and the IWSLT 2025 test set; for CS-to-EN, we use the IWSLT 2024 dev set~\cite{ahmad-etal-2024-findings} and IWSLT 2025 test set.

In \cref{tab:systems-overview}, we report system statistics that exclude incomplete logs (e.g., missing segments, mismatched predictions and delays, or severely hallucinated outputs).

\begin{table}[t]
\centering
    \resizebox{\linewidth}{!}{%
    \begin{tabular}{@{}lllcc@{}}
    \toprule
    \multirow{2}{*}{Regime} & \multirow{2}{*}{Lang. Pair} & \multirow{2}{*}{Dataset} & \multicolumn{2}{c}{Systems / Teams} \\
    & & & All & Filtering \\ \midrule
    \multirow{5}{*}{Short} & \multirow{3}{*}{EN$\rightarrow$DE} & IWSLT 22 test & 68 / 5 & 46 / 5 \\
    & & IWSLT 23 test & 5 / 5 & 4 / 4 \\
    & & tst-COMMON & 71 / 6 & 49 / 6 \\
    & EN$\rightarrow$JA & tst-COMMON & 9 / 3 & 7 / 3 \\
    & EN$\rightarrow$ZH & tst-COMMON & 14 / 3 & 7 / 3 \\ \midrule
    \multirow{4}{*}{Long} & EN$\rightarrow$DE & ACL 6060 dev / IWSLT 25 & 20 / 6, 10 / 6 & — \\
    & EN$\rightarrow$JA & ACL 6060 dev / IWSLT 25 & 16 / 3, 3 / 2 & — \\
    & EN$\rightarrow$ZH & ACL 6060 dev / IWSLT 25 & 16 / 4, 8 / 4 & — \\
    & CS$\rightarrow$EN & IWSLT 24 dev / IWSLT 25 & 14 / 2, 4 / 2 & — \\ \bottomrule
    \end{tabular}%
    }
\caption{Overview of systems in short- and long-form evaluations. ``All/Filtering'' refers to the number of systems before and after filtering degenerate systems (see \S\ref{sec:degenerated}; short-form only).}
\label{tab:systems-overview}
\end{table}

\subsection{Meta-Evaluation Settings}

\subsubsection{Metrics}
\paragraph{True Latency.}
\label{def:true-latency}
To enable fair comparisons across latency metrics, we require a reference latency reflecting the user experience, i.e., how long the user needs to wait for translation.
Since human evaluation is infeasible at scale, we adopt a carefully designed automatic approximation, which we refer to as \textit{true latency}.
This is grounded in an intuitive and practical definition of latency in speech translation:
\emph{On average, how long does a user have to wait for a given piece of source information to appear in the translation?}
Concretely, we define true latency as the average delay between each target word and its corresponding source word:
\begin{equation}
    TL = \frac{1}{\mathbf{|Y^A|}}\sum^{\mathbf{|Y^A|}}_{i=1}{d_i-d^{src}_i},
\end{equation}
where $d_i$ is the emission time of the target word $y_i$, and $d^{src}_i$ is the corresponding source delay.
We define the source delay as the time when the speaker finished the word if there is only one corresponding source word, or the last word corresponding to the target word if the target word aligns with multiple source words $
    d^{src}_i = \max_{l}{\{s^{end}_l|(y_i,s_l) \in \mathcal{A}(\mathbf{Y}\rightarrow\mathbf{S})\}}$,
where $s^{end}_l$ is the end timestamp of the source word $s_l$, and $\mathcal{A}(\mathbf{Y}\rightarrow\mathbf{S})$ is the translation alignment between the target and the source. 
As discussed in \S\ref{subsec:short}, computing the latency over all words, including tail words, can introduce systematic bias.
To mitigate this, we restrict the true latency calculation to words generated strictly during simultaneous decoding (i.e., before the end-of-source signal), and consider only the subset of target words $\mathbf{Y^A} \subseteq \mathbf{Y}$ aligned to at least one source word, thus avoiding biases introduced by over or under-generation~\cite{polak-etal-2022-cuni,papi-etal-2022-generation}. 
This raises a natural question: \emph{Why use automatic latency metrics if true latency is available?} In practice, true latency requires high-quality transcripts and reliable word-level alignments, which are often unavailable---especially for low-resource languages. It also involves multiple processing steps and is substantially more complex than standard metric evaluation. Importantly, as shown in \S\ref{subsec:short-eval} and \S\ref{subsec:long-eval}, automatic metrics can approximate true latency with high accuracy when best practices are followed, making them a practical alternative in most cases.

\paragraph{Score Difference.} For the main evaluation, we adopt the pairwise comparison approach \cite{mathur-etal-2020-tangled}.
Rather than independently evaluating each system, we examine the difference between the scores of two systems: $\Delta = score(\text{System A}) - score(\text{System B})$.
Pairwise comparison better reflects the typical use case of latency metrics---distinguishing between two systems.
More importantly, pairwise comparison allows us to focus on relative differences rather than absolute values.
We also restrict comparisons to system pairs evaluated on the same test set and language pair.

\paragraph{Accuracy.}\label{def:accuracy} Following \cite{kocmi-etal-2021-ship}, we evaluate the accuracy of binary comparisons: \emph{given a pair of systems, which is better according to the true latency ranking} (used as gold labels)?
Accuracy is defined as the proportion of system pairs whose relative ranking by the metric matches that of the true latency:
\begin{equation*}
    \text{Accuracy} = \frac{|sign(\Delta\text{TL}) = sign(\Delta \text{M})|}{|\text{all system pairs}|}.
\end{equation*}
Accuracy considers only the ranking, not magnitude, of the latency differences, allowing us to aggregate comparisons between languages and test sets with different scales.
We use bootstrap resampling with $N=10000$ \cite{tibshirani1993introduction} and consider all metrics within the 95\%

\subsubsection{Implementation}

Below, details on short- and long-form regimes are provided. 

\paragraph{Short-Form Regime.}
We tokenize hypotheses, reference transcripts and translations using \texttt{MosesTokenizer} (Chinese/Japanese with \texttt{KyTea}~\cite{neubig-mori-2010-word}).
We perform a time alignment between the source speech and golden transcripts using \gls{mfc}~\cite{mcauliffe17_interspeech} to obtain word timestamps.
We use \texttt{awesome-align}~\cite{dou-neubig-2021-word} with a finetuned \texttt{bert-base-multilingual-cased} model to map hypothesis words to source words.
Finally, we compute the true latency as~\S\ref{def:true-latency}, keeping only alignments with probability $>0.5$ and excluding punctuation.

\paragraph{Long-Form Regime.}
We follow the same true latency definition, but use WhisperX~\cite{bain23_interspeech} for alignment instead of \gls{mfc}, as it is more robust to challenging IWSLT~2025 conditions (restarts, repetitions, domain terminology, non-native speech).
We also resegment the system hypotheses before alignment, since \texttt{awesome-align} has a~512-token input limit.

\section{Short-Form Evaluation}
\label{subsec:short-eval}

\begin{figure*}[t]
    \centering
    \includegraphics[width=\linewidth]{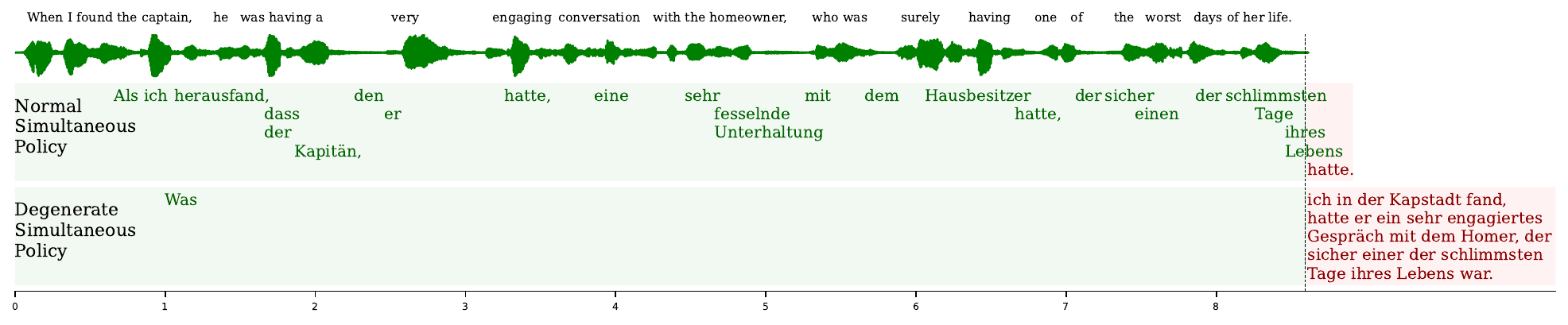}
    \caption{Translations and emission times of a \gls{sst} model.
    \textcolor{ForestGreen}{Green translations} are emitted simultaneously, while  \textcolor{Red}{red translations} are emitted after the end-of-segment signal (vertical dashed line).
     The \textbf{Normal Simultaneous Policy} emits the translations uniformly and has only a small fraction of tail words.
     The \textbf{Degenerate Simultaneous Policy} quickly emits a few words at the beginning, while waiting to translate the majority of words after the segment ends, effectively performing offline translation. 
     }
    \label{fig:degen}
\end{figure*}
\begin{figure}[t]
    \centering
    \includegraphics[width=\linewidth]{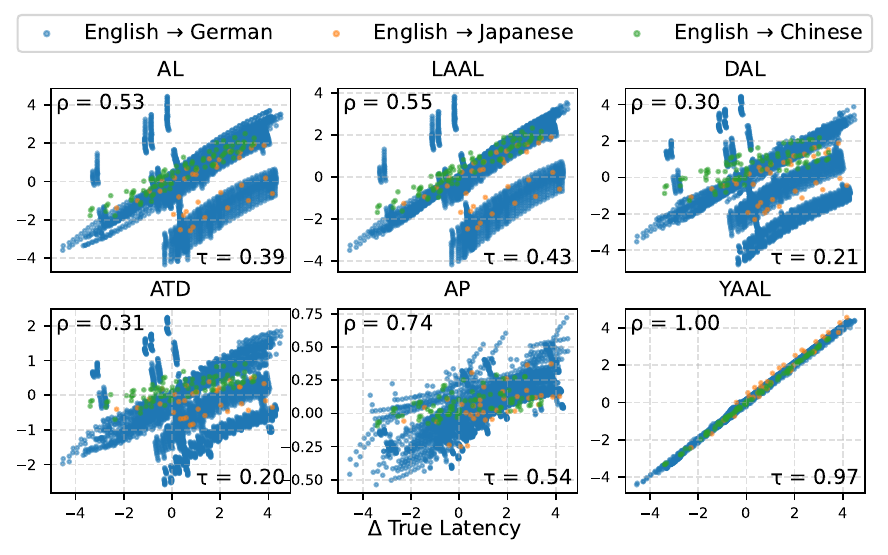}
    \caption{Each point represents the difference between the true latency (x-axis) and the automatic metric (y-axis) for two systems. Reported Pearson and Kendall rank correlations are indicative, as each language pair has a different scale.}
    \label{fig:short-form-corr}
\end{figure}

We present pairwise comparisons of all short-form systems in \cref{fig:short-form-corr}.
All metrics show a positive correlation between true and automatic latency.
However, some system pairs exhibit anomalous behavior---near-vertical lines far from the diagonal---particularly with AL, LAAL, DAL, and ATD.
These systems follow a \emph{degenerate simultaneous policy}: they emit a low-latency prefix but translate most of the sentence offline after the segment boundary (see example in \cref{fig:degen}).
Since all current latency metrics (AL, LAAL, DAL, ATD, AP) include tail words in their computation, they assign a high latency to these systems, even though the initial prefix is translated with low latency.
YAAL avoids this by only counting words translated strictly before the segment end, enabling detection of degenerate policies (described below).

\begin{figure}[t]
    \centering
    \includegraphics[width=\linewidth]{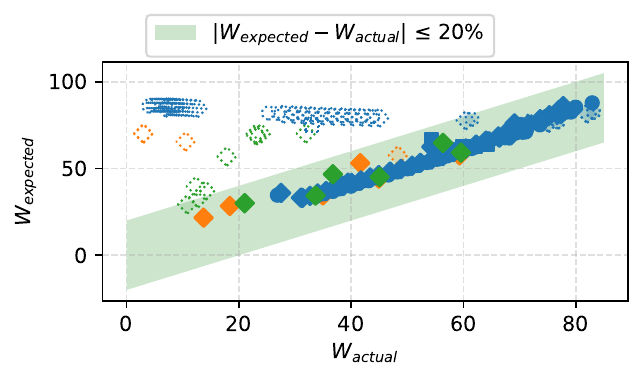}
    \caption{Proposed degenerate simultaneous policy test (\textcolor{ForestGreen}{green area}). Empty markers represent automatically filtered systems. Dotted markers are systems submitted by the teams whose systems were reported in \cref{tab:offenders}. Each point represents the actual and expected proportion of words translated simultaneously as observed on the short-form systems. Blue color represents En$\rightarrow$De, orange color En$\rightarrow$Ja, and green color En$\rightarrow$Zh systems, respectively. Diamonds present \texttt{tst-COMMON}, circles IWSLT 2022, and squares IWSLT 2023 test set systems.}
    \label{fig:test}
\end{figure}

\begin{table}[t]
    \centering
    \resizebox{\linewidth}{!}{%
    \addtolength{\tabcolsep}{-3pt}
    \begin{tabular}{llc|r|cccccc|r}
    \toprule
        Lang. & Test Set & Team & $\Delta$ & AL & LAAL & AP & DAL & ATD & YAAL & TL \\
        \midrule

        \multirow{3}{*}{En-De} & test22 & A & 81 & 4.9 & 4.9 & 0.92 & 5.8 & 3.6 & 0.8 & 0.9 \\
         & test23 & B & 42 & 1.9 & 2.0 & 0.82 & 3.8 & 2.5 & 1.5 & 1.5 \\
         & tst-C. & A & 78 & 4.1 & 4.1 & 0.88 & 5.2 & 3.2 & 0.8 & 0.9 \\
         \midrule
        En-Ja & tst-C. & A & 67 & 4.8 & 4.8 & 0.90 & 5.0 & 1.5 & 1.6 & 1.5 \\
         \midrule
        En-Zh & tst-C. & C & 46 & 2.2 & 2.3 & 0.99 & 3.8 & 1.1 & 1.6 & 1.5 \\
        
        \bottomrule
    \end{tabular}%
    }%
    \caption{
    Systems with the largest gaps between expected and actual fractions of simultaneous words. $\Delta = |W_{\text{expected}} - W_{\text{actual}}|$, TL = True Latency.
    }
    \label{tab:offenders}
\end{table}

\paragraph{Detecting Degenerate Simultaneous Policy.}
\label{sec:degenerated}
To detect degenerate simultaneous policies, we propose a simple test: a comparison of the observed and expected fractions of simultaneously translated words.
The observed fraction of words translated strictly simultaneously across the entire test set $\mathbf{S}$ can be easily computed from the observed delays and segment durations as:
\begin{equation}
    W_{\text{actual}} = \frac{\sum_s^{ |\mathbf{S}|}|\{ d_i^s | d_i^s < |{X}_s| \}|}{\sum_s^{ |\mathbf{S}|} |{Y}_s|}
\end{equation}

A normal system will likely translate words evenly across the segment; therefore, we can use the observed automatic latency $L_\text{YAAL}$ to estimate the fraction of words translated simultaneously.
If the average segment length is $X_\text{avg}$, then $X_\text{avg} - L_\text{YAAL}$ seconds of a segment will be translated simultaneously, while the remaining $L_\text{YAAL}$ seconds will contain words translated after the end of the segment signal, since the entire translation should be shifted by $L_\text{YAAL}$ seconds to the right.
The expected fraction of words translated simultaneously is then:
\begin{equation}
      W_{\text{expected}} = \frac{\sum_s^{ |\mathbf{S}|}\max(0,  |X_s| - L_\text{YAAL})}{\sum_s^{ |\mathbf{S}|} |{X}_s|}.
      \label{eq:expected}
\end{equation}
Note that some segments may be shorter than the observed latency $L_\text{YAAL}$; in such cases, we assume that no words were translated simultaneously in that segment (hence the use of the $\max(0, \cdot)$ function).

As discussed above, a degenerate simultaneous policy leads to fast initial emissions, followed by a long tail of offline translations (see~\cref{fig:degen}).
In such cases, most words are translated after the end of the segment, resulting in a low observed fraction of simultaneously translated words $W_{\text{actual}}$.
However, the latency of the prefix generated simultaneously remains low, resulting in a high expected fraction of words translated simultaneously $W_{\text{expected}}$.
Therefore, \emph{if the expected simultaneously-translated word fraction significantly exceeds the observed one, i.e., $W_{\text{expected}} \gg W_{\text{actual}}$, we can conclude that the system follows the degenerate simultaneous policy.}
Note that we rely on the YAAL metric for this comparison, as the other latency metrics include tail words in their computation.

In \cref{fig:test}, most of the systems exhibit a linear relationship between the expected and observed word proportions ($W_{\text{expected}}\sim W_{\text{actual}}$), while some show significantly higher expected than actual proportions.
We report the most extreme cases in \cref{tab:offenders}, where the differences range from 42\% to 81\%.
Automatic metrics (AL, LAAL, DAL, ATD, AP) predict much higher latency than the true latency, whereas YAAL matches predictions.
Filtering these outliers reveals a clear linear trend with maximum differences of only 14\%.
\emph{We classify systems with $|W_{\text{expected}} - W_{\text{actual}}| > 20\%$ as following the degenerate simultaneous policy.}

\begin{table}[!t]
\centering
\resizebox{\linewidth}{!}{%
\begin{tabular}{l|cccccc|r}
\toprule
$p$-val/Lang. & AL & LAAL & DAL & ATD & AP & YAAL & N \\
\midrule 
\multicolumn{8}{c}{all system pairs} \\
\midrule  
All & 0.64 & 0.67 & 0.57 & 0.54 & 0.73 & \textbf{0.98} & 4900 \\\midrule
\multicolumn{8}{c}{w/o degenerate simultaneous policy} \\\midrule
All & 0.96 & \underline{0.99} & 0.97 & 0.93 & 0.88 & \textbf{0.99} & 2259 \\\midrule
En-De & 0.96 & \underline{0.99} & 0.97 & 0.93 & 0.88 & \textbf{0.99} & 2217 \\
En-Ja & 0.95 & \textbf{1.00} & \textbf{1.00} & 0.90 & 0.86 & \textbf{1.00} & 21 \\
En-Zh & \textbf{1.00} & \textbf{1.00} & 0.95 & 0.95 & 0.81 & \textbf{1.00} & 21 \\\midrule
Different Team & 0.95 & \underline{0.98} & 0.96 & 0.91 & 0.82 & \textbf{0.98} & 1406 \\
Same Team & 0.98 & \underline{0.99} & 0.99 & 0.96 & 0.99 & \textbf{1.00} & 853 \\
\bottomrule
\end{tabular}%
}
\caption{Accuracy of short-form systems. Best scores in \textbf{bold}. \underline{Underlined} scores are considered tied with the best metric.}
\label{tab:accuracies-main-result}
\end{table}

\paragraph{Which is the best Short-form Latency Metric?}
Moving to metrics, we observe (see \cref{fig:short-form-corr}) that all show positive correlations with the true latency, but each language pair has a slightly different scale, which \emph{motivates the use of accuracy instead of a simple correlation} in \cref{tab:accuracies-main-result}.
If we consider all system pairs (including systems with a degenerate simultaneous policy), we see that all metrics significantly underperform YAAL (by more than 25\% absolute), which reaches an accuracy of 98\%.
If we remove systems with the degenerate simultaneous policy, all metrics gain a significant boost in accuracy (bottom part of \cref{tab:accuracies-main-result}). 
YAAL still remains the best metric, but LAAL becomes tied with it as the best-performing metric when considering all system pairs and across all language pairs.
The other metrics also improve, with AL and DAL reaching accuracies above 95\%.
However, ATD and AP still lag behind by a significant margin of 6\% and 11\% absolute, respectively.
This difference is reduced when we consider only pairs coming from the same team, but even in this case, YAAL and LAAL still outperform the other metrics.
Overall, we conclude that \emph{even though the latency metrics make simplifying assumptions, they approximate true latency with high accuracy.}
However, the presence of \emph{degenerate simultaneous policies can significantly affect the accuracy, and therefore, we recommend using YAAL together with the proposed test to ensure a reliable evaluation of short-form \gls{sst} systems.}

\begin{figure*}[t]  
    \centering
    \includegraphics[width=\linewidth]{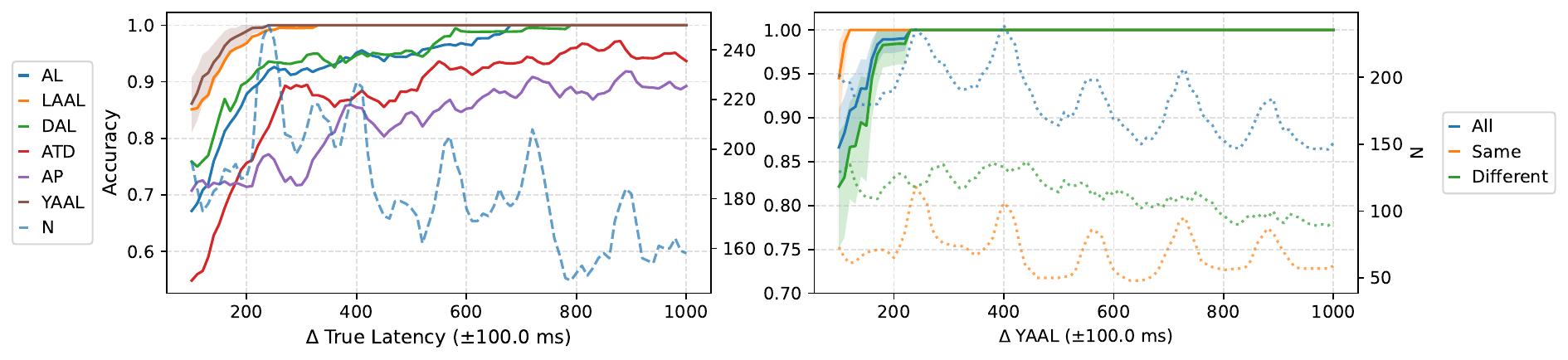}
    \caption{Short-form accuracies after removing degenerate simultaneous policy systems based on the difference in:  \textbf{left:} true latency, and 
    \textbf{right:} YAAL values of two systems.
     $N$ (dashed/dotted lines) indicate the number of pairs in each group. 
     The colored strips show the 95\% confidence interval. }
    \label{fig:short-form-acc}
\end{figure*}

\paragraph{How Sensitive are the Automatic Metrics?}
In the left part of \cref{fig:short-form-acc}, we plot the accuracy as a function of the true latency difference after removing degenerate systems.
Overall, the accuracies show a clear increasing trend with respect to the true latency difference.
YAAL outperforms all metrics, and LAAL is tied within the 95\% confidence interval.
ATD shows poor performance at small differences, but improves steeply between 100-300 ms, likely due to its fixed 300 ms word duration assumption.
However, ATD remains 5-10\% behind YAAL/LAAL at larger differences.
AP performs worst, reaching only 90\% accuracy at large differences with a consistent 10--15\% lag behind YAAL.
In the right part of \cref{fig:short-form-acc}, we analyze YAAL accuracy as a function of the difference in YAAL values between two systems.
These results provide practical guidance for interpreting YAAL differences: accuracy increases with larger YAAL differences, reaching 90\% accuracy at 40-240 ms difference and 99\% accuracy at 110-310 ms difference.

%
%
%
%
%
%
%
%
%
%
%
%
%
%
%
%
%
%
%
%
%
%
%
%
%
%

%
%
%
%

\paragraph{Should we use the Short-Form Regime?}
\label{par:should-we-use-shortform}
As discussed in \S\ref{sec:pitfalls}, short-form evaluation can distort latency assessment.
In \cref{fig:offline-words}, we show that a substantial fraction of translations occur \textit{after} the end-of-segment signal---from 41\% in low-latency regimes (1--2s) to 72\% in high-latency regimes (4--5s)---effectively translating offline.
\textit{Short-form evaluation with artificial boundaries and problematic tail-word handling often misrepresents \gls{sst} system behavior, underscoring the need for long-form evaluation} (\S\ref{subsec:long-eval}).

\begin{table}[t]
\centering
\footnotesize
\begin{tabular}{l|cccc}
\toprule
Latency regime [s] & 1--2 & 2--3 & 3--4 & 4--5 \\
\midrule
Tail Words [\%] & 41 & 49 & 63 & 72 \\
\bottomrule
\end{tabular}%
    \caption{
    Average fraction of words after the end-of-segment signal in short-form evaluation, across systems.
    }
    \label{fig:offline-words}
\end{table}

\section{Long-Form Evaluation}
\label{subsec:long-eval}

\paragraph{Which Resegmentation is Better?}
\label{subsub:resegmentation}
In \cref{tab:resegment-concat}, we evaluate two resegmentation tools: \mwerSegmenter{} and our proposed \segmenter{}.
The evaluation is conducted on reconcatenated short-form outputs, allowing us to compare with gold segment boundaries.
As we can see in \cref{tab:resegment-concat}, \textit{the accuracy of latency evaluation is significantly higher with the proposed segmenter}.
A limitation of this experiment is that the outputs are generated in short-form mode, which may not reflect the behavior of true long-form systems.
The results, however, strongly indicate that the proposed \segmenter{} is better suited for long-form latency evaluation.

%
%
%
%
%
%
%
%
%
%
%
%
%

\begin{table}[t]
\centering
\resizebox{\columnwidth}{!}{%
\begin{tabular}{@{}l|cc@{}}
\toprule
    & \mwerSegmenter{} & \segmenter{} \\ 
\midrule
Latency (StreamLAAL) & 86.4 & 94.0 \\
MT Quality (COMET) & 99.3 & 99.1 \\
\bottomrule
\end{tabular}%
}
\caption{Accuracy of latency and quality after resegmentation.}
\label{tab:resegment-concat}
\end{table}

\paragraph{Do we need Resegmentation?}
The upper part of \cref{tab:accuracies-longform} compares the original short-form metrics \emph{without resegmentation} on long-form systems.
We see that the accuracies are low, not exceeding 65\% on all systems. 
Compared to StreamLAAL, which employs resegmentation, the best-performing AL metric loses 17\% absolute, and the gap is even wider compared to the proposed LongYAAL, with AL falling short by 29 points. 
The lower part of \cref{tab:accuracies-longform} reports the accuracy of latency metrics in long-form systems \emph{with resegmentation}. 
We see that the resegmentation quality significantly influences the accuracy.
StreamLAAL and LongLAAL share the same definition but differ only in the resegmentation---while StreamLAAL uses the \mwerSegmenter{}, LongLAAL (and all other ``Long-'' metrics) uses our \segmenter{}.
The gap in accuracy is 12\% points, showing a trend similar to the one in \cref{tab:resegment-concat}.

We hypothesize the significant improvement with resegmentation stems from metrics (except AP) computing latency as a difference from an ``oracle'' policy assuming uniform word durations, no silences, and monotonic alignment.
With longer inputs, these assumptions no longer hold, leading to misaligned observed and expected policies.
Resegmentation restores validity by aligning outputs to reference segments, aligning the conditions with the short-form evaluation, and ensuring the oracle policy assumptions hold.
\textit{These results underscore the critical role of segmentation in reliable long-form latency evaluation.}

\begin{table}[!t]
\centering
\resizebox{\columnwidth}{!}{%
\begin{tabular}{l|ccccccc|r}
\toprule
\multicolumn{9}{c}{longform + unsegmented} \\
\midrule 
$p$-val/Lang. &  & AL & LAAL & DAL & ATD & AP & YAAL & N \\
\midrule 
All &  & \textbf{0.65} & \underline{0.62} & 0.57 & \underline{0.61} & 0.39 & \underline{0.62} & 594 \\
\toprule 
\multicolumn{9}{c}{longform + resegmented} \\
\midrule 
$p$-val/Lang. & \multicolumn{1}{c|}{\Stream{LAAL}} & \Long{AL} & \Long{LAAL} & \Long{DAL} & \Long{ATD} & \Long{AP} & \Long{YAAL} & N \\
\midrule 
All & \multicolumn{1}{c|}{0.82} & 0.92 & \textbf{0.94} & \textbf{0.94} & \underline{0.93} & 0.71 & \textbf{0.94} & 594 \\\midrule
En-De & \multicolumn{1}{c|}{\underline{0.92}} & \underline{0.95} & \textbf{0.95} & \underline{0.94} & \textbf{0.95} & \underline{0.93} & \underline{0.94} & 226 \\
En-Zh & \multicolumn{1}{c|}{0.86} & \underline{0.92} & \textbf{0.95} & \underline{0.94} & \underline{0.93} & 0.55 & \textbf{0.95} & 148 \\
En-Ja & \multicolumn{1}{c|}{0.63} & 0.85 & \underline{0.89} & \textbf{0.92} & 0.85 & 0.43 & \textbf{0.92} & 123 \\
Cs-En & \multicolumn{1}{c|}{0.77} & \underline{0.97} & \underline{0.98} & \textbf{0.99} & 0.96 & 0.80 & \underline{0.98} & 97 \\\midrule
Different Team & \multicolumn{1}{c|}{0.82} & 0.90 & \underline{0.92} & \textbf{0.93} & \underline{0.90} & 0.70 & \underline{0.92} & 413 \\
Same Team & \multicolumn{1}{c|}{0.83} & 0.99 & \textbf{1.00} & 0.98 & 0.98 & 0.73 & 0.99 & 181 \\
\bottomrule
\end{tabular}%
}
\caption{Accuracy of long-form systems. Best scores in \textbf{bold}. \underline{Underlined} scores are considered tied. Metrics in the bottom half use the proposed \segmenter{}, except for StreamLAAL that uses the original \mwerSegmenter{}.}
\label{tab:accuracies-longform}
\end{table}

\begin{figure*}[!ht]  
    \centering
    \includegraphics[width= \linewidth]{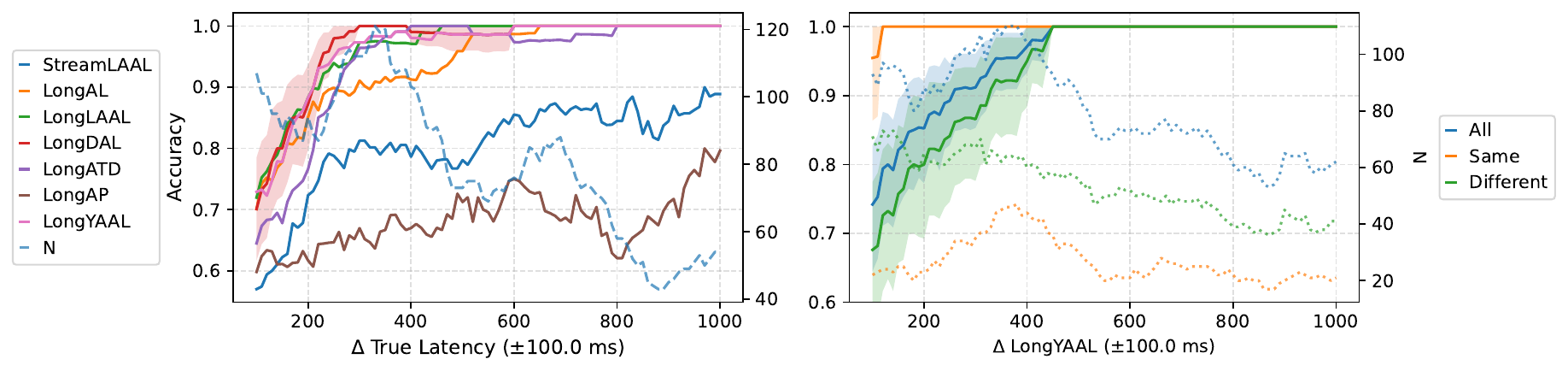}
    \caption{Long-form accuracies based on the difference in:  \textbf{left:} true latency of two systems, and 
    \textbf{right:} LongYAAL values of two systems.
     $N$ (dashed/dotted lines) indicate the number of pairs in each group. 
     The colored strips show the 95\% confidence interval. }
    \label{fig:long-form-acc}
\end{figure*}

\paragraph{Which is the best Long-form Latency Metric?} 
The bottom part of \cref{tab:accuracies-longform} shows accuracy results for long-form metrics with resegmentation.
The best-performing metrics are LongYAAL, LongLAAL, and LongDAL, all achieving accuracies above 93\% and tied within the 95\% confidence interval.
LongATD and LongAL also perform well, exceeding 92\%.
StreamLAAL, using the \mwerSegmenter{}, significantly underperforms all other long-form metrics with only 82\% accuracy, confirming the limitations of its resegmentation approach.
LongAP shows the lowest accuracy at 71\%, likely due to sensitivity to variable segment lengths.
Overall, we conclude that \textit{LongYAAL, LongLAAL, and LongDAL are the best-performing long-form latency metrics}.
Yet, we recommend LongYAAL as it considers all words in the output without any modifications, potentially offering better generalization across test sets.

\paragraph{How Sensitive are the Long-Form Automatic Metrics?}
In the left portion of \cref{fig:long-form-acc}, we plot the accuracy of long-form metrics as a function of the true latency difference between two systems.
Overall, the accuracies show a clear increasing trend with respect to the true latency difference, similar to the trends observed in short-form metrics.
LongYAAL, LongLAAL, and LongDAL show similar performance, with LongYAAL lagging behind at intermediate differences (250--600 ms).
We hypothesize that this is due to the fact that LongYAAL considers all words in the output, while LongLAAL and LongDAL modify the output by removing tail words (LongLAAL) or modifying the observed delays (LongDAL), which may make LongYAAL more sensitive to segmentation errors around the segment boundaries.
However, these differences are marginal (1--2\% absolute) and may not be observed in other test sets.
LongATD first lags behind at small differences but then shows a steep increase, likely due to its fixed 300 ms word duration assumption.
StreamLAAL shows a significantly lower accuracy than all other long-form metrics, reaching only 90\% accuracy at a 1000 ms difference, confirming the limitations of its resegmentation approach.
Finally, LongAP shows the lowest accuracy at all latency differences.
In the right portion of \cref{fig:long-form-acc}, we analyze the accuracy of LongYAAL as a function of the difference in LongYAAL values between two systems.
We see that the accuracy increases with the difference in LongYAAL values, reaching 90\% accuracy at around a 260 ms difference and nearly perfect accuracy at around a 440 ms difference, similar to the trends observed in short-form YAAL.

\section{Conclusions}
\label{subsec:conclusion}

In this paper, we present the first comprehensive evaluation of latency metrics for both short-form and long-form \gls{sst}.
Our key finding is that \emph{automatic latency metrics}---despite their strong assumptions about uniform word timings, absence of silences, and monotonic alignments---can \emph{reliably approximate actual latency} when applied according to best practices.
However, we also demonstrate that short-form evaluation can be misleading due to artificial input segmentation, which inadvertently incentivizes degenerate simultaneous policies that misalign with actual user experience.
To address this limitation, we introduce YAAL for short-form evaluation and propose a simple diagnostic test based on expected versus observed proportions of simultaneous words to detect such degenerate policies.
Nevertheless, even with YAAL, short-form evaluation has inherent limitations stemming from artificial segmentation: up to 70\% of words are generated after the segment ends, a condition unrepresentative of simultaneous translation.
We therefore recommend prioritizing long-form evaluation whenever feasible, as it better reflects real-world usage and avoids segmentation artifacts.
For long-form \gls{sst}, we show that existing short-form metrics cannot be directly applied without resegmentation.
To address this challenge, we introduce LongYAAL, extending YAAL to long-form settings with a novel resegmentation method \segmenter{}, which overcomes limitations of the standard approach.
LongYAAL achieves superior performance compared to the existing long-form metric StreamLAAL, with performance nearly matching YAAL in short-form settings---a strong indicator of viable latency assessment for long-form \gls{sst}. The code is available in \toolkit{}.\footnote{\url{\toolkiturl}}

\paragraph{Limitations.}
Although our study offers a comprehensive evaluation of the latency metrics for \gls{sst} and introduces improved tools for both short- and long-form regimes, some limitations remain.
First, our experimental analysis focuses on systems from the IWSLT Shared Tasks, which may not fully represent the range of techniques or data conditions used in broader academic or industrial settings.
Second, our analysis focuses on high-resource languages for which data were available, but the findings should be reconfirmed in low-resource language settings.
Third, although the proposed \segmenter{} improves alignment robustness, word-level alignment is still susceptible to errors in the presence of disfluencies or noise.

\paragraph{Future Work.} We argue that some form of resegmentation or semantic alignment will likely remain necessary for the foreseeable future, as all existing latency metrics rely on the ``oracle'' policy, which assumes uniform word timings, no silences, and monotonic alignments---assumptions that become less valid with longer inputs.
Resegmentation helps restore these assumptions to some extent.
While our proposed resegmentation method improves significantly thanks to \mwerSegmenter{}, there is still room for further enhancement.
Another promising direction is to explore alternative evaluation paradigms that do not rely on the ``oracle'' policy assumptions, potentially eliminating the need for resegmentation altogether. 

\section{Generative AI Use Disclosure}

We used generative AI tools solely for correcting grammar and polishing the manuscript, but they were not used for producing the manuscript, and they are not co-authors of this paper.

\bibliographystyle{IEEEtran}


\end{document}